\documentclass[conference]{IEEEtran}

\usepackage{csquotes}
\MakeOuterQuote{"}

\usepackage{microtype}
\usepackage{cite}
\usepackage{amsmath,amssymb,amsthm}
\usepackage{amsfonts}
\usepackage{graphicx}

\usepackage{subcaption} 
\usepackage{enumitem} 
\usepackage[pagebackref=true,breaklinks=true,colorlinks,bookmarks=false]{hyperref}

\def\Vec#1{{\boldsymbol{#1}}} 

\begin{document}

\title{Buy Me That Look: An Approach for Recommending Similar Fashion Products}

\author{
\IEEEauthorblockN{Abhinav Ravi}
\IEEEauthorblockA{\textit{Data Sciences} \\
\textit{Myntra}\\
Bengaluru, India \\
abhinav.ravi@myntra.com}
\and
\IEEEauthorblockN{Sandeep Repakula}
\IEEEauthorblockA{\textit{Data Sciences} \\
\textit{Myntra}\\
Bengaluru, India \\
sandeep.r@myntra.com}
\and
\IEEEauthorblockN{Ujjal Kr Dutta}
\IEEEauthorblockA{\textit{Data Sciences} \\
\textit{Myntra}\\
Bengaluru, India \\
ujjal.dutta@myntra.com}
\and
\IEEEauthorblockN{Maulik Parmar}
\IEEEauthorblockA{\textit{Data Sciences} \\
\textit{Myntra}\\
Bengaluru, India \\
parmar.m@myntra.com}

}

\maketitle

\begin{abstract}
Have you ever looked at an Instagram model, or a model in a fashion e-commerce web-page, and thought \textit{"Wish I could get a list of fashion items similar to the ones worn by the model!"}. This is what we address in this paper, where we propose a novel computer vision based technique called \textbf{ShopLook} to address the challenging problem of recommending similar fashion products. The proposed method has been evaluated at Myntra (www.myntra.com), a leading online fashion e-commerce platform. In particular, given a user query and the corresponding Product Display Page (PDP) against the query, the goal of our method is to recommend similar fashion products corresponding to the entire set of fashion articles worn by a model in the PDP full-shot image (the one showing the entire model from head to toe). The novelty and strength of our method lies in its capability to recommend similar articles for all the fashion items worn by the model, in addition to the primary article corresponding to the query. This is not only important to promote cross-sells for boosting revenue, but also for improving customer experience and engagement. In addition, our approach is also capable of recommending similar products for User Generated Content (UGC), eg., fashion article images uploaded by users. Formally, our proposed method consists of the following components (in the same order): i) Human keypoint detection, ii) Pose classification, iii) Article localisation and object detection, along with active learning feedback, and iv) Triplet network based image embedding model.
\end{abstract}

\section{Introduction}

The surge in online shopping due to the proliferation of numerous fashion e-commerce platforms like Zalando (\url{zalando.com}), Ssence (\url{www.ssense.com/en-us}), Myntra (\url{www.myntra.com}), and Farfetch (\url{www.farfetch.com}), added to the fact that fashion products top across all categories in online retail sales \cite{jagadeesh2014large}, necessitates the importance of efficient and effective fashion product recommendations. However, contrary to traditional product recommendations, fashion product recommendation is challenging. This is because fashion products are often displayed under various settings (\textit{eg.}, clean in-shop clothes, clothes worn by a model with studio/ street background etc), having enormous amount of variations present in the fashion items (\textit{eg.}, color, texture, shapes, viewpoint, illumination and styles).

\begin{figure}[t]
  \centering
  \includegraphics[width=\linewidth]{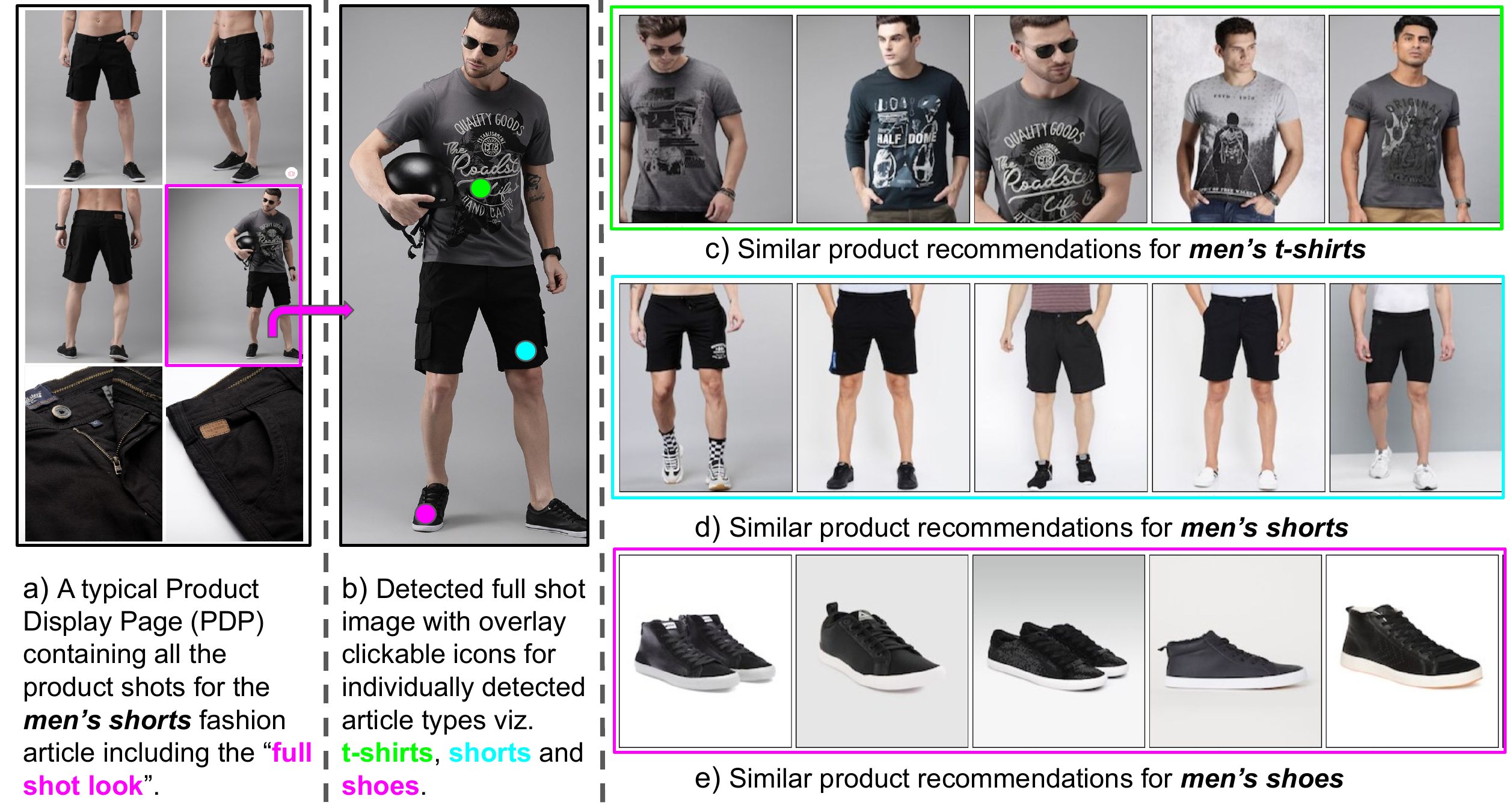}
  \caption{An illustration of the problem of \textit{similar products recommendation}. (a) A given Product Display Page (PDP) for the article type \textit{men's shorts}, (b) Identification of the full-shot image among all the PDP images in (a), identifying different products/ article types in that image (\textit{t-shirt} with green overlay icon, \textit{men's short} with cyan overlay icon, and \textit{shoe} with pink overlay icon), and (c-e) Retrieval of similar products as present in the full-shot image in (b). Note that color of the bounding boxes around retrieved images in (c-e) correspond to the color of the respective overlay icons in (b).}
  \label{fig_teaser}
\end{figure}

Myntra (www.myntra.com) is an Indian fashion e-commerce platform that hosts a large-scale collection of fashion and lifestyle items. For a given user query, the platform displays a Product Display Page (PDP) containing the relevant fashion item. As shown in Figure \ref{fig_teaser}, the PDP consists of different \textit{views}, \textit{looks} or \textit{shots} of the product. A \textit{full-shot look} image in the PDP refers to the one that is displaying a model from the head to the toe (Figure \ref{fig_teaser}-b). As observed, the model in this full-shot look image not only wears the primary product for which the query is made (for eg, \textit{men's short}), but also a few other secondary products, like \textit{t-shirt}, \textit{shoes}) etc. Often the user might be interested in buying the entire \textit{look} of the model, i.e., Shopping the entire model Look (\textbf{ShopLook}!). This use case is also relevant to the Instagram-like feature on the Myntra platform where \textit{fashion influencers} regularly post images, and users might be willing to \textit{mimic} the looks of their \textit{influencers}.
\begin{figure}[t]
\centering
	\includegraphics[width=\linewidth]{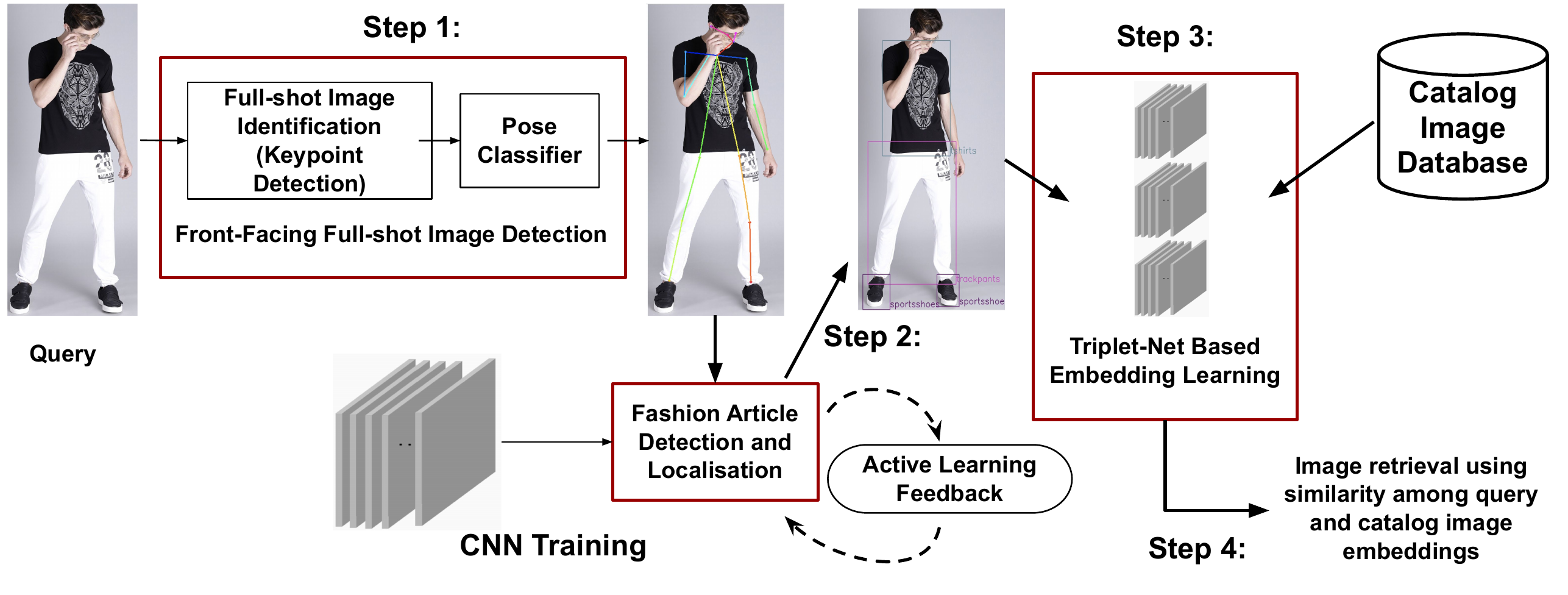}
    \caption{An illustration of the pipeline of the proposed framework.}
    \label{framework_shoplooks}
\end{figure}

This problem can be addressed by retrieving and recommending similar products from our database. However, given the highly occluded images in our platform (for example, with the model arm, other complementary objects etc), along with the huge pose variations, this is a challenging problem. Additionally, in contrast to performing recommendation for a single, primary article (for the query), we need to recommend similar products for the entire set of fashion articles worn by a model. This is an important problem because the additional recommendation of similar products for the secondary articles could generate further revenue by promoting cross-sells, and also improve customer experience and engagement. The nature of our platform (for instance, type of garments, poses, etc) makes it impossible to address this problem by training an existing Artificial Intelligence (AI) / Machine Learning (ML) model on a public benchmark dataset. Moreover, popular datasets like DeepFashion2 \cite{DeepFashion2}  are available for non-commercial research purposes only. For this reason, we propose an approach consisting of state-of-the-art Computer Vision (CV) based components, and train it on our own large scale collection of real-world fashion products.

A high-level illustration of our method can be seen in Figure \ref{framework_shoplooks}. Given a user query, the Product Display Page (PDP) shows images of the product in different views or angles, including the \textit{full-shot look} image. To automatically identify the full-shot look image from the entire set of PDP images, we first perform human key-point detection to verify the presence of head and ankle keypoints. We then make use of a pose classifier to obtain a \textit{front-facing} full-shot image. This image is used to perform article localization and fashion object detection. This step is further enhanced by a human-in-the-loop active learning like feedback. Corresponding to a fashion item, we build image similarity model using embedding learning. The obtained embeddings are used to compute similarity with respect to objects in the database, and recommend the most similar ones. Our large-scale end-to-end approach being generic in nature, can also be utilized in other multimedia applications, like product recommendations in social media, apart from fashion e-commerce platforms.

\section{Background and Related Work}
\label{sec_relwork}

In this section, we shall briefly discuss some of the related works that provide a background for our approach: i) Human Keypoint Estimation, ii) Object Detection, and iii) Embedding learning. We shall also provide a brief categorization of different fashion applications.

\textbf{Human Keypoint Estimation:}
To identify a \textit{full-shot look} image among all the PDP images, we make use of a heuristic criterion leveraging a \textit{state-of-the-art} computer vision based human key-points estimation technique by Xiao \textit{et al.} \cite{xiao2018simple}. Other relevant pose estimation methods include: i) Cascaded Pyramid Network (CPN) \cite{chen2018cascaded} (dominant on the COCO 2017 key-point challenge), ii) Hourglass method \cite{newell2016stacked} (dominant on the MPII benchmark), and iii) CMU-Pose \cite{cao2017realtime} (bottom-up approach that makes use of Part Affinity Fields). The method by Xiao \textit{et al.} outperforms other competitive approaches despite being much simpler in nature.

\textbf{Object Detection:}
Darknet architecture based single-stage detectors like YOLO \cite{redmon2018yolov3} have been the choice among researchers for the task of real-time object detection. However, our goal of object detection do not require real-time output, as this component of our pipeline can be done offline. Hence, we would prefer to pick a model with a better mean Average Precision (mAP) score, while disregarding the run-time latency. The Mask RCNN \cite{he2017mask} method has been chosen for this purpose.

\textbf{Embedding learning:}
Embedding learning \cite{veit2017conditional,schroff2015facenet} seeks to learn representations of raw images such that similar examples are grouped together, while moving away dissimilar ones. To retrieve images of catalog database fashion products that are similar to the products present in a query image, we make use of embedding learning to obtain representations, and compute image similarity. The image similarity between a pair of image embeddings $\Vec{x}_i,\Vec{x}_j \in \mathbb{R}^d$ can be either computed using a \textit{cosine similarity}: $\textrm{cos}(\Vec{x}_i,\Vec{x}_j)=\frac{\Vec{x}_i^\top\Vec{x}_j}{||\Vec{x}_i||.||\Vec{x}_j||}$ or a \textit{squared Euclidean distance}: $\delta^2(\Vec{x}_i,\Vec{x}_j)=(\Vec{x}_i-\Vec{x}_j)^\top(\Vec{x}_i-\Vec{x}_j)$.

\textbf{Fashion Applications:}
Fashion applications can be broadly categorized from three standpoints: 1. Retrieval of similar clothing items based on a query fashion item \cite{ak2018learning}, 2. The extraction of attributes (\textit{eg.}, texture, colour, pattern etc) from a given fashion item \cite{luo2018coarse}, and 3. Recommendation of complementary clothes, given a query fashion item \cite{li2020graph,yang2019interpretable,yang2019transnfcm}. Kalantidis \textit{et al.} \cite{kalantidis2013getting} also attempted at retrieving multiple fashion items in an image, similar to ours. However, their clustering based object detection technique performs poorly on images with complex backgrounds.

\section{Proposed Method}
\label{sec_proposed_method}

\begin{figure}[t]
\centering
	\includegraphics[width=0.8\columnwidth]{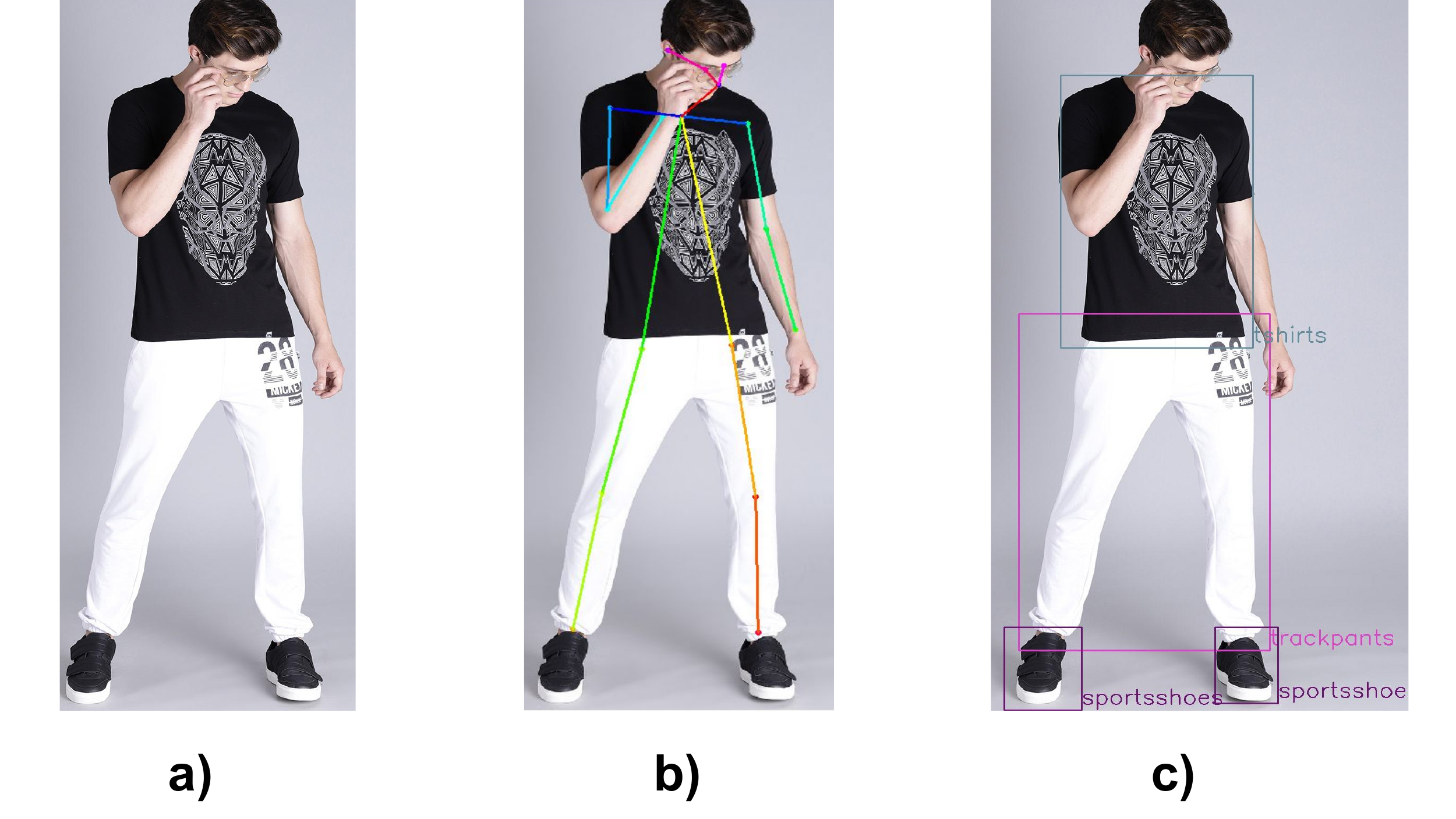}
    \caption{a) A sample catalog full-shot image, b) Illustration of the obtained keypoints for the person in a), and c) Article type detection and localisation for a full-shot image.}
    \label{KP-objDet}
\end{figure}
\begin{figure}[t]
\centering
	\includegraphics[width=0.7\columnwidth]{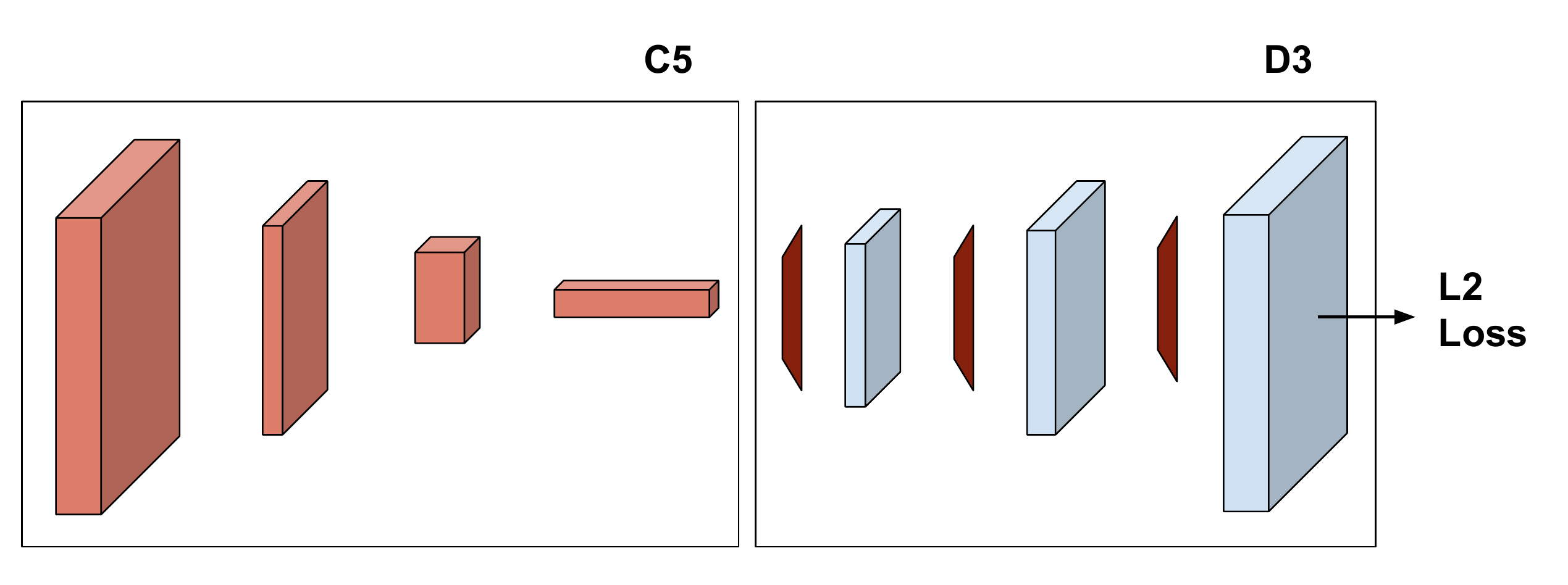}
    \caption{Architecture of the human key-point detection component in our method.}
    \label{architecture_KP-objDet}
\end{figure}

\begin{figure*}[t]
\centering
    \begin{subfigure}{0.55\linewidth}
        \centering
        \includegraphics[width=\linewidth]{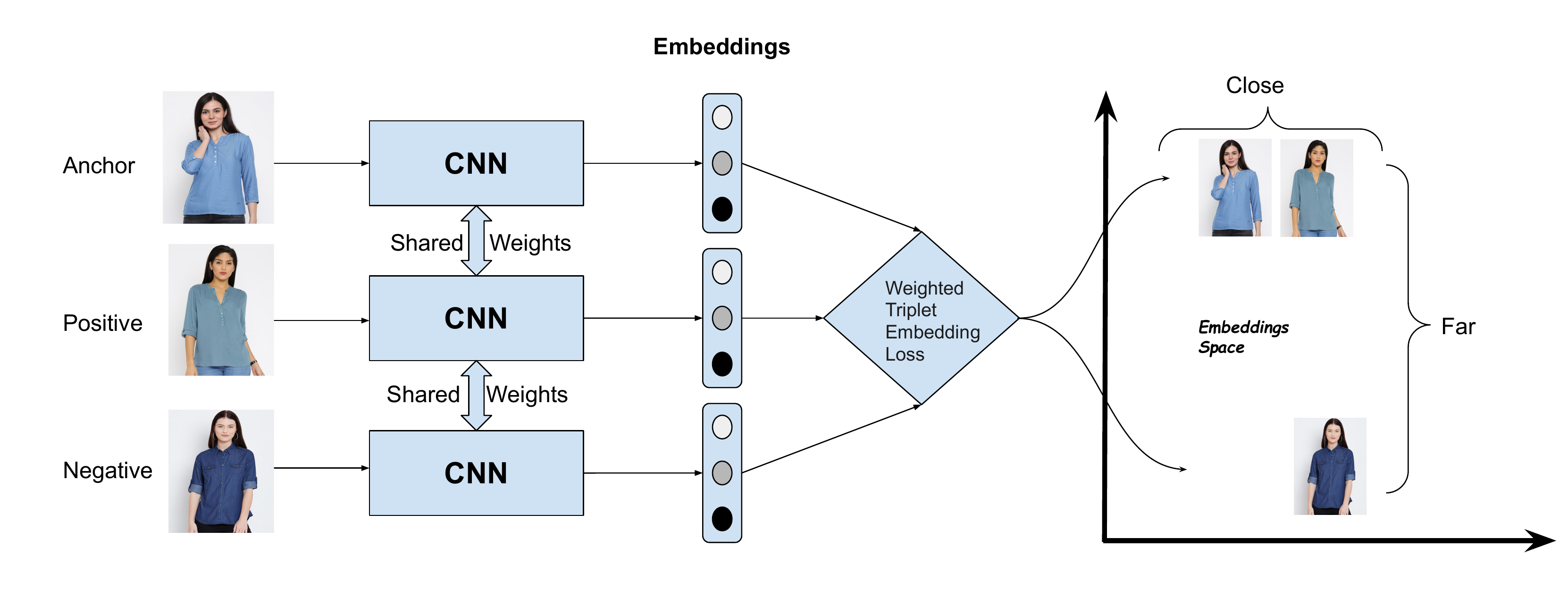}  
        \caption{}
        \label{Triplet}
    \end{subfigure}
    \begin{subfigure}{0.35\linewidth}
        \centering
    	\includegraphics[width=\linewidth]{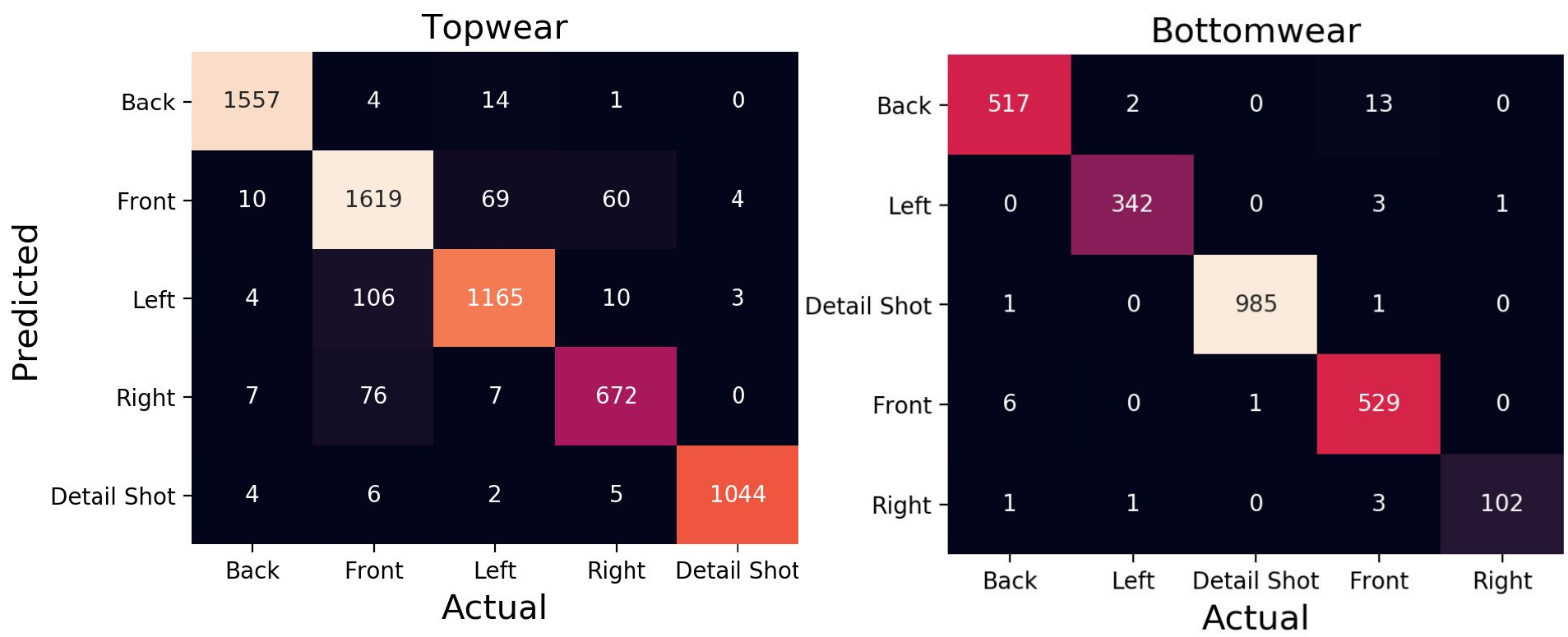}
    	\caption{}
        \label{poseclassif_confmat}
    \end{subfigure}
    \caption{(a) An illustration of the triplet network used in our approach for obtaining embeddings to compute image similarity. We made use of ResNets as our backbone CNN, while employing a \textit{weighted triplet embedding loss}. (b) Confusion matrices corresponding to two broad article types, obtained by the pose classifier.}
    \label{trip_confmat}
\end{figure*}

We now discuss our proposed method. It consists of the following major steps:
\begin{enumerate}
    \item \textbf{Front-facing Full-shot Image Detection}
    
    Given a user query, the Product Display Page (PDP) shows images of the product in different views or angles, including the \textit{full-shot look} image. To automatically identify the full-shot look image from the entire set of PDP images, we first perform human key-point detection. To do so, we use the technique proposed by Xiao \textit{et al.} \cite{xiao2018simple}, as mentioned earlier. It has a ResNet \cite{ResNet} backbone to perform feature extraction. After the last convolution stage of the ResNet (C5 in Figure \ref{architecture_KP-objDet}), three deconvolutional layers (D3 in Figure \ref{architecture_KP-objDet}) are added (with batch normalization and ReLU). There are 256 filters in each layer (with $4\times4$ kernel, and stride of 2). The predicted heatmaps $\{H_1,\cdots,H_k \}$ for $k$ key-points are generated using a $1\times 1$ convolutional layer at the end. $L_2$ loss is applied between the target and predicted heatmaps. Having obtained the key-points, we verify the presence of head and ankle key-points, to identify our full-shot look image.
    
    Among the full-shot look images, there may be images where a model may be facing towards left, right, or even backwards. Such images may provide only an occluded view of the articles of interest. Therefore, as shown in Step 1 of Figure \ref{framework_shoplooks}, we add another sub-component for performing pose classification of the image into one of the following categories: front, back, left, right or detailed shot. In our work, we made use of a supervised ResNet18 network as our classifier, the annotations of which are performed by our in-house taggers. For a sample \textit{front-facing full-shot look} catalog image in Figure \ref{KP-objDet}-a, the obtained key-points are shown in Figure \ref{KP-objDet}-b.
    \item \textbf{Fashion article Detection and Localisation}
    
    Step 2 of Figure \ref{framework_shoplooks} shows the next stage of our method. The \textit{front-facing full-shot look} image obtained from the previous step contains multiple fashion articles and accessories worn by the model. Corresponding to each of these articles, we have to recommend a list of similar fashion products. For this subtask of identifying different article types, we must crop, or segment out the individual Regions Of Interests (ROIs) from the \textit{full-shot look image}. However, a mere segmentation of the fashion article might render out incomplete article information due to occlusion from other peripheral objects. Hence, we train the article type detection and localisation module using the \textit{bounding box} tags for the fashion articles present in these images. Table \ref{tab:articleTable} lists the targeted fashion articles present in our training images.
    
    For the fashion article detection and localisation task, we train the Mask RCNN model \cite{he2017mask} on a custom training dataset, which gives us the bounding box location, and classification for around 20 apparel types as mentioned in Table\ref{tab:articleTable}. Figure \ref{KP-objDet}-c shows the detected article types for the full-shot look image in consideration.

    Additionally, we incorporate an active learning setting, by employing our in-house taggers to identify misclassified examples, and make use of them for re-training the model. As shown in the experiments, this leads to further gains in performance metrics.
    \item \textbf{Embedding generation for article types}
    
    Having extracted the relevant fashion articles from the full-shot look image, we now need to retrieve similar fashion products from the catalog database. For this, we seek to represent the extracted article types, and the products from the database, in a common embedding space that groups together similar articles while moving away dissimilar ones. For this, we make use of a triplet based network architecture to learn our embeddings, as illustrated in Figure \ref{Triplet}. A triplet network consists of three identical Convolutional Neural Networks (CNN) with shared weights (each of which may be regarded as a \textit{branch}). In our case, we experimented with different configurations of ResNet, as shown later. To train it, one requires triplets of images such that the first two of which are sematically similar, while the third being dissimilar to the first two. Let, the embeddings for the triplet of images be denoted as $(\Vec{x}_a, \Vec{x}_p, \Vec{x}_n)$. Here, $\Vec{x}_a$ is called as the query, or \textit{anchor}, $\Vec{x}_p$ is called as the \textit{positive}, and $\Vec{x}_n$ is called as the \textit{negative}. The objective for training the network is to bring the embeddings $\Vec{x}_a$ and $\Vec{x}_p$ closer, while moving away $\Vec{x}_n$.
    
    This is achieved by minimizing the following \textit{weighted triplet loss}, defined as:
    \begin{equation}
    \label{equation:l_total}
    \mathcal{L}_{total} =  \mathcal{L}_{triplet} + \alpha \mathcal{L}_{embedd}
    \end{equation}
    Here, $\mathcal{L}_{triplet}$ is the \textit{triplet margin ranking loss} \cite{veit2017conditional}, defined as: 
    \begin{equation}
    \label{equation:tripletmarginrankingloss}
    \mathcal{L}_{triplet} = max(0, m + \delta^2(\Vec{x}_a,\Vec{x}_p) - \delta^2(\Vec{x}_a,\Vec{x}_n) ),
    \end{equation}
    such that $\delta^2(\Vec{x}_i,\Vec{x}_j)=\left \| \Vec{x}_i -\Vec{x}_j \right \|_2^2 $ denotes the squared Euclidean distance between the pair of examples $\Vec{x}_i$ and $\Vec{x}_j$, with $\left \| \Vec{x}_i \right \|_2^2$ being the squared $l_2$ norm of $\Vec{x}_i$. $\alpha>0$ is a trade-off hyper-parameter in (\ref{equation:l_total}). The objective of (\ref{equation:tripletmarginrankingloss}) is to constrain the squared Euclidean distance of the anchor-negative pair to be larger than the squared Euclidean distance of the anchor-positive pair by a margin $m>0$.
    
    Furthermore, the loss term $\mathcal{L}_{embedd}$ in (\ref{equation:l_total}) denotes the \textit{embedding loss}, and is defined as follows:
    \begin{equation}
    \label{equation:embeddloss}
    \mathcal{L}_{embedd} = \tau (\left \| \Vec{x}_a \right \|_2^2+ \left \| \Vec{x}_p \right \|_2^2 + \left \| \Vec{x}_n \right \|_2^2 ).
    \end{equation}
    Here, $\tau=\frac{1}{3d}$, such that $\Vec{x}_a,\Vec{x}_p,\Vec{x}_n \in \mathbb{R}^d$, \textit{i.e.}, $d$ is the embedding size. Essentially, $\mathcal{L}_{embedd}$ performs a normalization of the representations of the examples in the triplet to ensure that the image embeddings remain within the radius range of the margin value. As the three CNNs (or \textit{branches}) of the triplet network share weights among themselves, we may pass a raw image through any of the branches, and obtain an embedding for computing similarity using either of cosine similarity or Euclidean distance (as discussed in the related work). Based on the the computed image similarity between the embedding of a query fashion article and that of a database product, we can retrieve the set of most similar products for the query.
\end{enumerate}

\begin{table}[t]
\centering
\caption{Targeted fashion article categories}
\label{tab:articleTable}
\resizebox{0.6\columnwidth}{!}{%
\begin{tabular}{|c|c|}
\hline
\textbf{Broad article category} & \textbf{Finer article types} \\ \hline
Topwear & Women tops, Shirts, T-shirts \\ \hline
Outerwear & \begin{tabular}[c]{@{}c@{}}Sweaters, SweatShirts, Jackets, \\ Blazers, Shrug, NehruJackets\end{tabular} \\ \hline
BottomWear & \begin{tabular}[c]{@{}c@{}}Jeans, Trousers, Shorts, Track pants,\\  Palazzos, Capris\end{tabular} \\ \hline
Skirts & Skirts \\ \hline
Dresses & Women dress \\ \hline
Footwear & Sports shoes, Casual shoes \\ \hline
Bags & Hand bags \\ \hline
\end{tabular}
}
\end{table}

\section{Experimental Results}
\label{sec_emp_studies}

\textbf{Front-facing Full-shot Image Detection Results}
As our initial experiment, we showcase the performance of Step 1 of our method shown in Figure \ref{framework_shoplooks}. As already mentioned, we make use of a ResNet18 network as our pose classifier to classify an image into one of the following categories: front, back, left, right or detailed shot. The training is done on an annotated dataset from our Myntra Catalog, accumulated by our in-house taggers. Figure \ref{poseclassif_confmat} shows the confusion matrices for two broad article types, the precision/recall for each of the categories are respectively: i) Top-wear: 98.7/ 98.4, 91.8/ 89.4, 90.4/ 92.6, 88.1/ 89.8 and 98.4/ 99.3, ii) Bottom-wear: 97.1/ 98.4, 98.8/ 99.1, 99.8/ 99.9, 98.6/ 96.3, and 95.3/ 99.0.
\begin{figure*}[t]
\centering
	\includegraphics[width=0.9\linewidth]{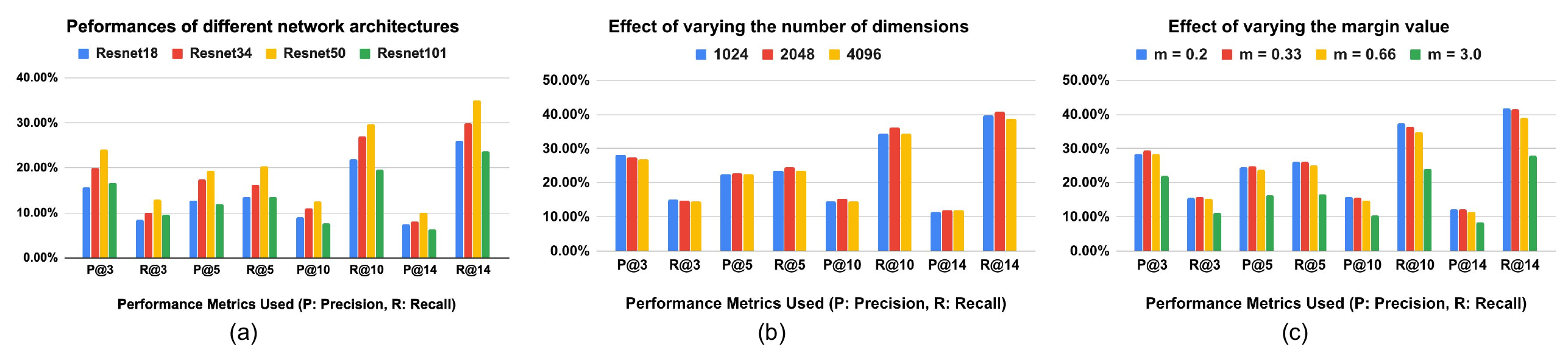}
    \caption{Additional studies for our image similarity model: (a) Effect of different ResNet architectures as the backbone CNN, (b) Effect of varying embedding sizes, and (c) Effect of varying the margin in the triplet loss.}
    \label{imgSimExpGraph}
\end{figure*}

\textbf{Fashion article Detection and Localisation results}
We now perform experiments to study the step 2 of our method as shown in Figure \ref{framework_shoplooks}. We train the Mask RCNN on a custom training dataset to provide us the bounding box location and classification from across the 20 apparel types, as mentioned in Table\ref{tab:articleTable}. The annotated training dataset for the Mask RCNN is obtained from our catalog, and consists of roughly 7-9k training images and around 800 test images for each fashion article type, resulting in a total of roughly 150k training images. We obtain an average mAP of 78\% for all the classes, while reaching as high as 92\% for some of the topwear classes (shirts and t-shirts). These values are inline with this model's performance on the Microsoft Common Objects in Context (MS COCO) data set \cite{MS_COCO}, which is around 60.3\% at IOU 0.5. This is justified because the COCO dataset has more number of classes with natural real world images, whereas images in our case are from 20 article categories. A sample object detection is shown in Figure \ref{KP-objDet}-c.

We use the "ResNet-101-FPN" variant for training, on a Tesla v100-PCIE-16GB GPU, with an image batch size of 16. It converges around 200k iterations. The standard hyper-parameters are taken from the paper (weight decay of 0.0001 and momentum of 0.9). The bounding-box (b-box) detection is taken as positive, only on a correct classification of bounding box for a IOU of 0.5 and above with ground truth b-box. The learning rate was initially kept at 0.03 and later reduced by a factor of 10 after 100k iterations. During inference, it takes roughly 400-500 ms to detect around 3-5 fashion objects in a detected full-shot look image.

We further employ our taggers to identify the misclassified examples, for retraining our model in an active learning setting. Table \ref{tab:activeLearning} reports the comparison of class wise Average Precision (AP), with and without active learning. As observed, the active learning component leads to an improved performance. 
\begin{table}[t]
\centering
\caption{Class wise AP improvement after employing active learning for some broad fashion categories.}
\label{tab:activeLearning}
\resizebox{0.6\columnwidth}{!}{%
\begin{tabular}{|c|c|c|}
\hline
\textbf{\begin{tabular}[c]{@{}c@{}}Broad \\ Category\end{tabular}} & \textbf{\begin{tabular}[c]{@{}c@{}} AP (in \%) without\\ active learning\end{tabular}} & \textbf{\begin{tabular}[c]{@{}c@{}} AP (in \%) with \\ active learning\end{tabular}} \\ \hline
topwear & 82.65 & \textbf{87.60} \\ \hline
bottomwear & 87.11 & \textbf{90.25} \\ \hline
outerwear & 81.35 & \textbf{83.98} \\ \hline
dress & 80.26 & \textbf{85.51} \\ \hline
skirts & 51.42 & \textbf{69.09} \\ \hline
footwear & 87.32 & \textbf{88.47} \\ \hline
bags & 69.71 & \textbf{78.37} \\ \hline
\end{tabular}
}
\end{table}

\textbf{Results on embedding generation for article types}
To obtain the embeddings for computing image similarity, we train a ResNet based triplet network for each of the broad article type in Table \ref{tab:articleTable}. To form triplets, the first image is used from the street2shop dataset \cite{hadi2015buy} which contains a garment item worn by a person in an uncontrolled setting (\textit{eg.}, streets, complex backgrounds). The second image contains the same garment object from our catalog. The pairs obtained in this manner help us in forming \textit{hard anchor-positive} pairs that are semantically similar, but have huge variations among them. A negative can be randomly sampled, containing an image from a different garment for same article type. Using such triplets in our method makes it robust to variations in the query image, and also lets us perform similar products recommendations from images in the wild, as shown later. We further perform \textit{semi-hard triplet mining} \cite{schroff2015facenet}, and retrain our model. 


We use ADAM optimizer with learning rate of $5e^{-5}$ and batch size of 32 on a Tesla v100-PCIE-16GB GPU. The value of $\alpha$ in equation \ref{equation:l_total} was fixed at $5e^{-5}$ following Veit \textit{et al.} \cite{veit2017conditional}. The training accuracies range from 92\% for bottomwear, to 98\% for topwear articles. The Precision (P) and Recall (R) values at different K values were used as the metric to evaluate our method quantitatively (using the ground-truth labels present with us).

Figure \ref{imgSimExpGraph} shows results of additional studies performed by changing the base architecture, embedding size, and margin in the embedding learning component of our method. \textbf{Our focus is to get a higher value of recall} (which we observed for $K=14$). Hence, \textbf{with respect to R@14}, the optimal set of hyperparameters for our method are: margin $m=0.2$ in (\ref{equation:tripletmarginrankingloss}) and embedding size $d=2048$ in (\ref{equation:embeddloss}). We also evaluated our method with different variants of the ResNet architecture, and observed a consistently better performance with ResNet50. For inference, we used the cosine similarity among embeddings.
\begin{figure}[t]
\centering
	\includegraphics[width=0.6\linewidth]{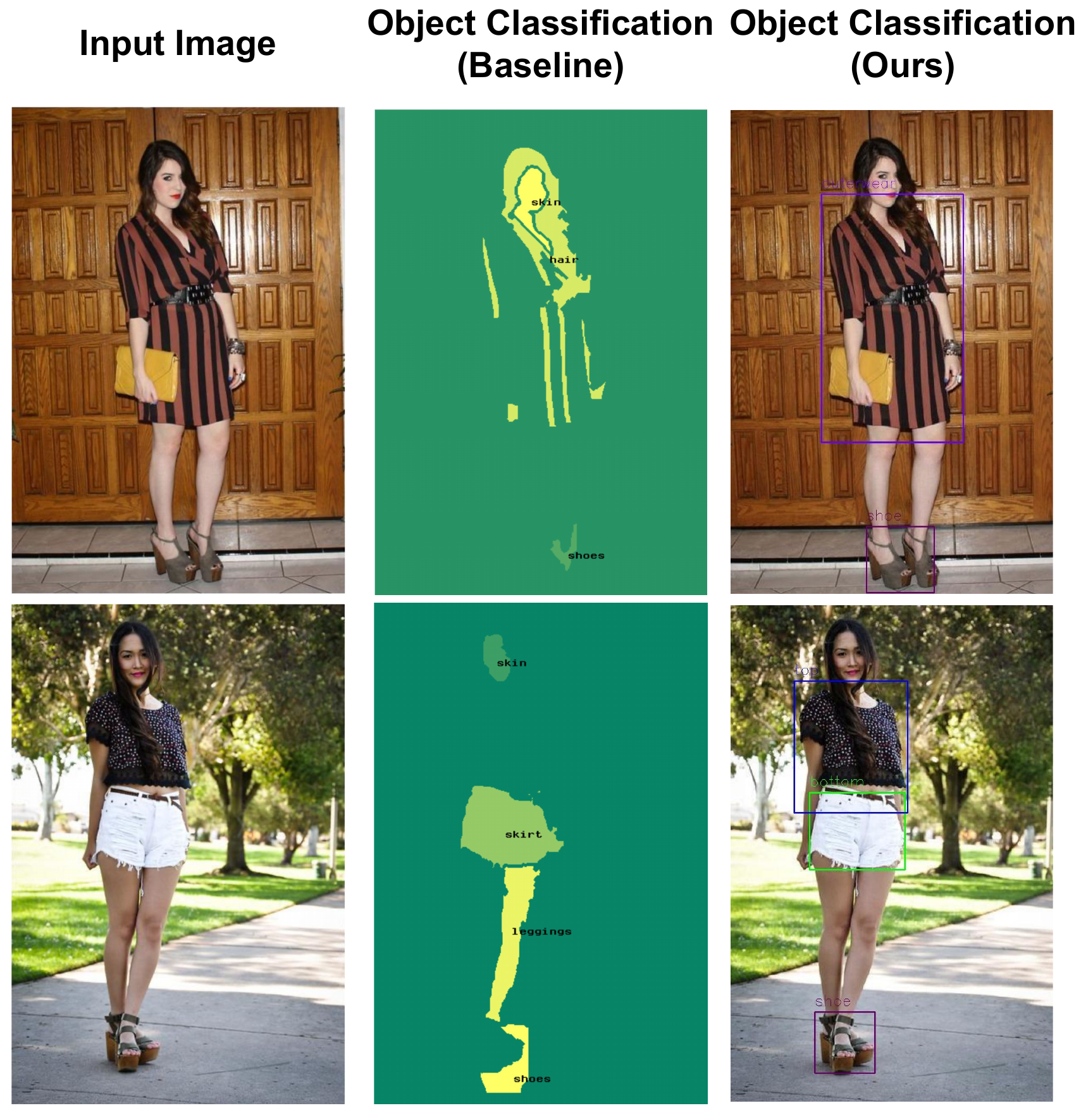}
    \caption{Qualitative comparison for the article type detection task. For each row, the left-most image is the input image. The middle column shows the article detection obtained by the baseline approach (they perform clustering to detect articles). The right-most images show the article detection results obtained by our method. We clearly obtain better qualitative results than our baseline.}
    \label{quali_comparisons}
\end{figure}

\begin{figure}[t]
\centering
	\includegraphics[width=0.6\columnwidth]{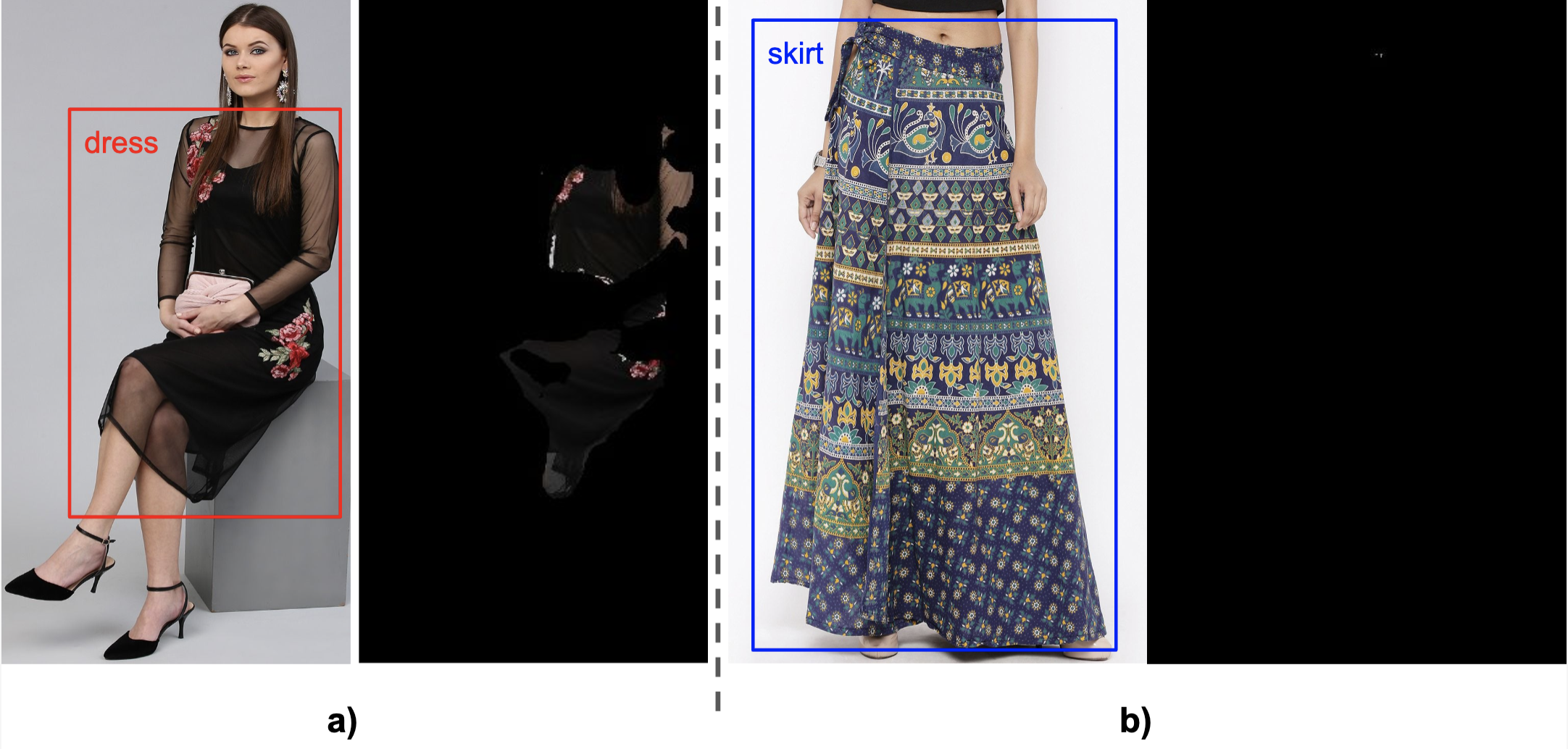}
    \caption{For both (a) and (b), the left image shows the detection by our bounding box based component, and the right ones show segmentation based object detection. As seen in (b), for an apparel with complex texture, the segmentation based method fails.}
    \label{seg_vs_bbox}
\end{figure}

\begin{figure}[t]
\centering
\begin{subfigure}{0.7\columnwidth}
    	\centering
		\includegraphics[width=\linewidth]{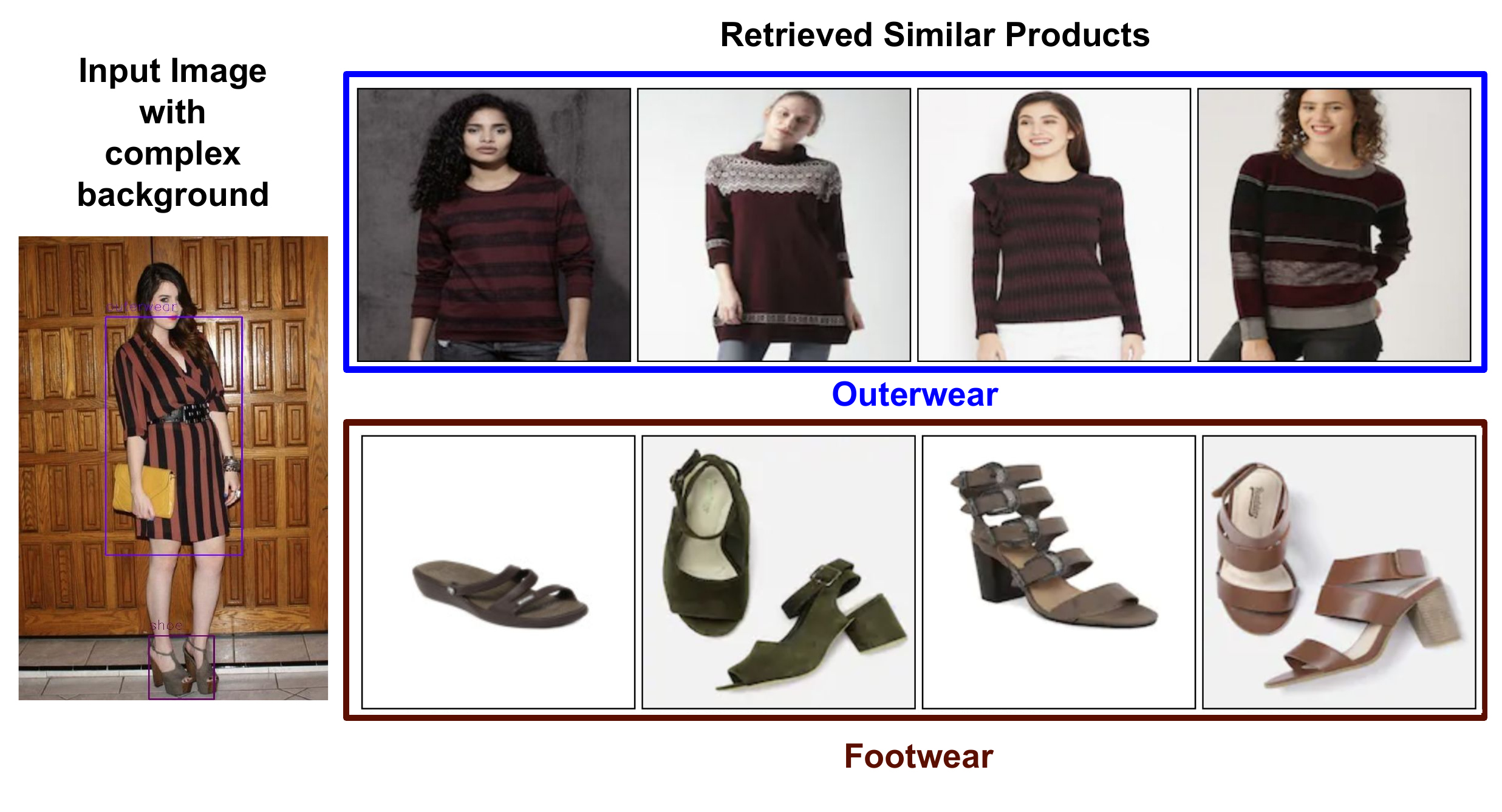}
\end{subfigure}\\
\begin{subfigure}{0.7\columnwidth}
    	\centering
		\includegraphics[width=\linewidth]{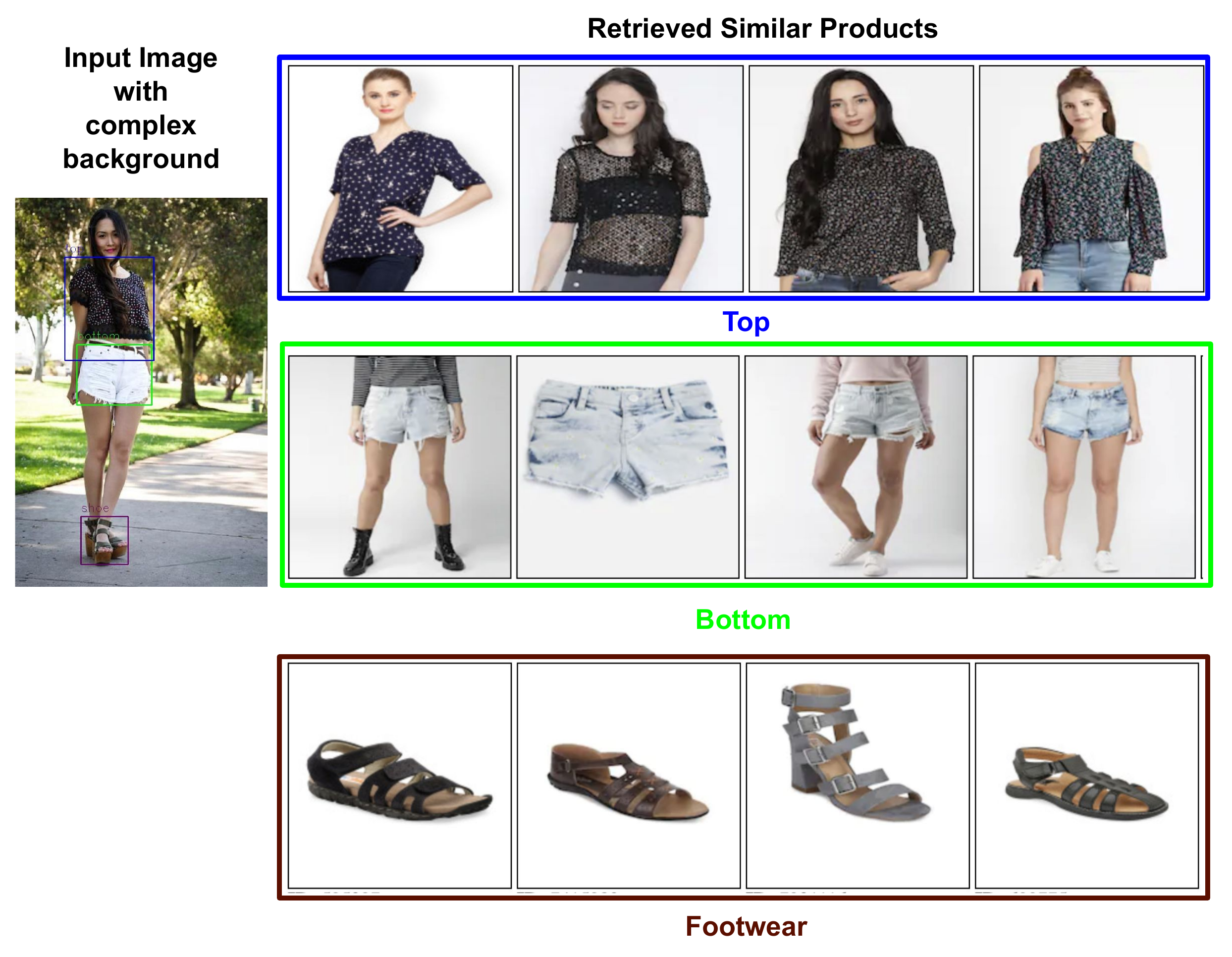}
\end{subfigure}\\
\caption{Qualitative results of retrieval of our method from source images with complex backgrounds.}
\label{retrieval_ours}
\end{figure}

\textbf{Qualitative comparison against baseline approaches}
We now compare our method qualitatively against two competing methods: i) The state-of-the-art method by Liang \textit{et al.} \cite{liang2018look} that performs segmentation to parse images, instead of bounding box based object detection used in step 2 of our method (shown in Figure \ref{framework_shoplooks}), and ii) The method by Kalantidis \textit{et al.} \cite{kalantidis2013getting} that makes use of a clustering based object detection technique. Figure \ref{seg_vs_bbox} shows that compared to the segmentation based approach by Liang \textit{et al.}, our component performs better. In Figure \ref{quali_comparisons}, we show two images with complex backgrounds, where the clustering based object detection used by Kalantidis \textit{et al.} fails to detect objects. On the other hand, our bounding box based component detects the products fairly well.
In Figure \ref{retrieval_ours}, we present the qualitative results of our end-to-end pipeline, on the two images used in Figure \ref{quali_comparisons}. We observe that our method is capable of recommending similar products fairly well. Retrieval results on a catalog image is also shown in Figure \ref{fig_teaser}.

\textbf{Quantitative comparison against baseline approaches}
We also compare the end-to-end quantitative performance of our method against a few end-to-end baselines: i) \textbf{SIFT}: A pipeline similar to ours, but by replacing the Triplet Net based embeddings by SIFT features, as in Kalantidis \textit{et al.} \cite{kalantidis2013getting}. We have already shown that our object detection component outperforms their clustering based object detection component. ii) \textbf{EUC}: A pipeline similar to ours, except the fact that Euclidean distance is computed between the embeddings during inference. iii) \textbf{CE}: A pipeline similar to ours, except the fact that a combination of cosine similarity and Euclidean distance is computed between the embeddings during inference. In Table \ref{vs_baselines_quant}, we show that our method outperforms all the baselines, thus justifying the choices made in our end-to-end pipeline.
\begin{table}[t]
\centering
\caption{Quantitative comparison of our approach against a few end-to-end baselines (all metrics are in $\%$).}
\label{vs_baselines_quant}
\resizebox{0.9\columnwidth}{!}{%
\begin{tabular}{|c|cc|cc|cc|cc|}
\hline
\textbf{Method}                                               & \textbf{P@3}   & \textbf{R@3}   & \textbf{P@5}   & \textbf{R@5}   & \textbf{P@10}  & \textbf{R@10}  & \textbf{P@14}  & \textbf{R@14}  \\ \hline
\begin{tabular}[c]{@{}c@{}} \textbf{SIFT}  \end{tabular}                                                   &0.9	&0.9	&0.7	&0.7	&0.5	&1.2	&0.4	&1.2          \\ \hline
\begin{tabular}[c]{@{}c@{}} \textbf{EUC}  \end{tabular}                                                   &24.0	&13.0	&19.3	&20.4	&12.5	&29.7	&10.0	&35.0          \\ \hline
\begin{tabular}[c]{@{}c@{}} \textbf{CE} \end{tabular} &26.3	&13.9	&21.0	&22.5	&13.5	&32.2	&11.0	&37.9          \\ \hline
\textbf{Ours}                                                 & \textbf{28.4} & \textbf{15.5} & \textbf{24.6} & \textbf{26.1} & \textbf{15.8} & \textbf{37.6} & \textbf{12.2} & \textbf{41.7} \\ \hline
\end{tabular}
}
\end{table}

\begin{figure}[t]
    \centering
    \includegraphics[width=\linewidth]{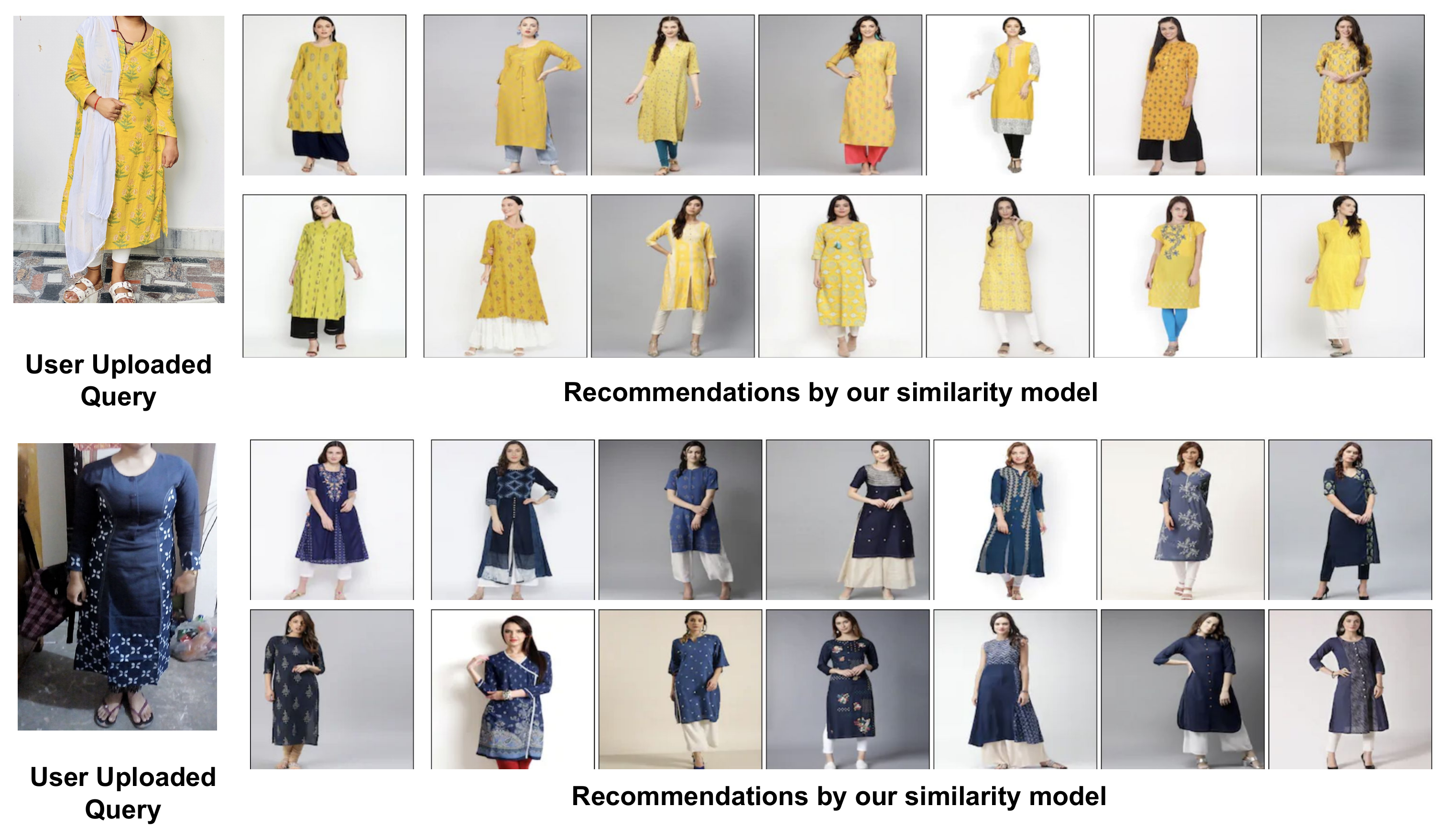}
    \caption{Retrieval performance of our image similarity component on challenging UGC images (Best viewed when zoomed-in).}
    \label{ugc_results}
\end{figure}

\textbf{Retrieval results beyond catalog images}: Apart from the Myntra catalog images, we also tested the performance of the image similarity component of our method on User Generated Contents (UGC), i.e., images of fashion objects uploaded directly by the users on our platform. Such UGC images usually have poorer resolution and/ or illumination, thus making effective recommendation a further challenging task. Figure \ref{ugc_results} shows a few retrieval results using the image similarity component of our method, on two query UGC images of the challenging ethnic wear category of \textit{Kurtas}. We observed fairly good qualitative retrieval results obtained by our method.

\textbf{Inference time of different components}:
Table \ref{tab:inferenceTime} reports the average inference times of different components of our method on a Tesla v100-PCIE-16GB GPU.
\begin{table}[t]
\centering
\caption{Inference time of different components}
\label{tab:inferenceTime}
\resizebox{0.45\columnwidth}{!}{%
\begin{tabular}{|c|c|}
\hline
\textbf{Model} & \textbf{\begin{tabular}[c]{@{}c@{}} Inference\\ time on GPU\end{tabular}} \\ \hline
Human pose & $\sim$200ms \\ \hline
Bounding Box & $\sim$400ms \\ \hline
Image Similarity & $\sim$100ms \\ \hline
Similarity Retrieval & $\sim$5ms \\ \hline
\end{tabular}
}
\end{table}

\textbf{Results of A/B testing and scope of future improvement}:
Despite the qualitative and quantitative improvements observed in our experiments, we wanted to evaluate our method by employing an online A/B experiment for our recommendation framework. Notably, we observed an increase in Click Through Rate (CTR) by 25\%, and add to cart ratio by 4\%. Furthermore, we also observed an increase in overall user session time, which depicts an improved customer engagement.

To give a better perspective, we divide the set of users in two halves: The first set for which we do not provide recommendations, and the second set, where we make recommendations based on our approach. We then compute specific metrics for each set of users. On average, we observed better metrics for the subset of users for which we make recommendations using our method. For example, we observed better \textit{add-to-cart} numbers, and an increase in the quantity of products selected.

Additionally, we employed a team of catalog experts to provide feedback on our retrieval results. Out of 500 randomly drawn similar products recommendation results, 477 were marked as visually relevant and correct (around 95\%). The experts provided us further feedback that we should take the following into account for a future version of our model: i) the occasion of an apparel/accessory (\textit{eg.}, workout, formal, party etc), and ii) the finer attributes (\textit{eg.}, neck type, sleeve length etc). Incorporation of these additional features in our model would further enhance the performance of our model. However, doing so is beyond the scope of this paper, and hence left as a future work.


\section{Conclusion}
\label{sec_conclusion}
This paper proposed a convenient and efficient method for automatic product searches, facilitating users to easily look for similar products as displayed on a product display page. We introduced a method that includes identifying the full-shot look of a product among the set of product display images, by identification of human key-points, followed by detection of broad fashion objects, and identification of similar products for each of the detected fashion articles by leveraging a triplet based embedding network. In the future, we would like to incorporate article attributes and occasion based filtering to facilitate better and robust product search.

\section*{Acknowledgment}
We are grateful to our manager Dr Ravindra Babu Tallamraju for his support and valuable feedback, and in being a source of inspiration and encouragement throughout the project.

\bibliography{shoplooks_MIPR21}
\bibliographystyle{unsrt}

\end{document}